\documentclass[10pt,twocolumn,letterpaper]{article}

\usepackage{cvpr}              % To produce the CAMERA-READY version
\usepackage{graphicx}
\usepackage{amsmath}
\usepackage{amssymb}
\usepackage{booktabs}
\usepackage[T1]{fontenc}    % use 8-bit T1 fonts
\usepackage{url}            % simple URL typesetting
\usepackage{booktabs}       % professional-quality tables
\usepackage{amsfonts}       % blackboard math symbols
\usepackage{nicefrac}       % compact symbols for 1/2, etc.
\usepackage{microtype}      % microtypography
\usepackage{xcolor}         % colors
\usepackage{multirow}
\usepackage{pifont}
\usepackage{float}
\usepackage[ruled]{algorithm2e}
\usepackage[pagebackref,breaklinks,colorlinks]{hyperref}

\usepackage[capitalize]{cleveref}
\crefname{section}{Sec.}{Secs.}
\Crefname{section}{Section}{Sections}
\Crefname{table}{Table}{Tables}
\crefname{table}{Tab.}{Tabs.}

\def\cvprPaperID{2267} % *** Enter the CVPR Paper ID here
\def\confName{CVPR}
\def\confYear{2022}

\begin{document}

\title{Instance-Dependent Label-Noise Learning with Manifold-Regularized Transition Matrix Estimation}
%

%\author{De Cheng$^{1}$, Tongliang Liu$^{2}$, Yixiong Ning$^{1}$, Nannan Wang$^{1}$\footnotemark[1], Bo Han$^{3}$,Gang Niu$^{4}$,\\ Xinbo Gao$^{5}$, Masashi Sugiyama$^{4}$ \\
%$^{1}$ Xidian University,
%$^{2}$ University of Sydney,
%$^{3}$ Hong Kong Baptist University, \\
%$^{4}$ RIKEN,
%$^{5}$ Chongqing University of Posts and Telecommunications.
%
%{\tt\small \{peilianghuang2017, junweihan2010, zhangdingwen2006yyy\}@gmail.com, dcheng@xidian.edu.cn}
%
%}
%
%\maketitle

\author{De Cheng$^{1}$, Tongliang Liu$^{2}$, Yixiong Ning$^{1}$, Nannan Wang$^{1}$\footnotemark[1], Bo Han$^{3}$, Gang Niu$^{4}$,\\ Xinbo Gao$^{5}$, Masashi Sugiyama$^{4,6}$. \\
$^{1}$ Xidian University,
$^{2}$ TML Lab, The University of Sydney,
$^{3}$ Hong Kong Baptist University, \\
$^{4}$ RIKEN,
$^{5}$ Chongqing University of Posts and Telecommunications,
$^{6}$ The University of Tokyo.\\
{\tt\small \{dcheng,nnwang\}@xidian.edu.cn, tongliang.liu@sydney.edu.au, yxning@stu.xidian.edu.cn,
}\\
{\tt\small  bhanml@comp.hkbu.edu.hk, gang.niu.ml@gmail.com,gaoxb@cqupt.edu.cn, sugi@k.u-tokyo.ac.jp
}
}

\maketitle

\footnotetext[1]{Corresponding author.}

%%%%%%%%% ABSTRACT
\begin{abstract}
   In label-noise learning, estimating the transition matrix has attracted more and more attention as the matrix plays an important role in building statistically consistent classifiers. However, it is very challenging to estimate the transition matrix $T(\textbf{x})$, where $\textbf{x}$ denotes the instance, because it is unidentifiable under the instance-dependent noise (IDN). To address this problem, we have noticed that, there are psychological and physiological evidences showing that we humans are more likely to annotate instances of similar appearances to the same classes, and thus poor-quality or ambiguous instances of similar appearances are easier to be mislabeled to the correlated or same noisy classes. Therefore, we propose assumption on the \emph{geometry} of $T(\textbf{x})$ that ``\emph{the closer two instances are, the more similar their corresponding transition matrices should be}''.
   More specifically, we formulate above assumption into the manifold embedding, to effectively reduce the degree of freedom of $T(\textbf{x})$ and make it stably estimable in practice. The proposed manifold-regularized technique works by directly reducing the estimation error without hurting the approximation error about the estimation problem of $T(\textbf{x})$.
   Experimental evaluations on four synthetic and two real-world datasets demonstrate that our method is superior to state-of-the-art approaches for label-noise learning under the challenging IDN.
   %Code is available at \url{http://github.com/noiselabellearning/manifold}.

   %creatively formulate the above observations into the manifold embedding framework to accurately approximate the instance-dependent transition matrix for noisy label learning.

   %Traditionally, the transition matrix learned from the estimated clean labels to noise labels has been widely exploited to optimize a clean label classifier assisting by the noisy data.
   %In this paper, we focus on estimating the instance-dependent label-noise transition matrix. We are the first to propose the hypothesis that ``the closer the distance between two instances, the more similar of their corresponding transition matrix'', then we creatively formulate the proposed hypothesis into the manifold embedding framework to estimate the instance-dependent transition matrix for noisy label learning. Extensive experiments on various datasets illustrate the effectiveness of our method which proves that the proposed hypothesis is very useful for the instance-dependent noisy label learning. Besides, we demonstrate better generalization and superior classification performances on both synthetic instance-dependent label noise datasets (MNIST, CIFAR10, CIFAR100) and the real-world human noise datasets (Clothing1M and Food101). Code is available at \url{http://github.com/noiselabellearning/manifold}.
\end{abstract}
 \vspace{-5mm}
%%%%%%%%% BODY TEXT
\section{Introduction}
\label{sec:intro}
%The remarkable success of deep neural networks(DNN) in representation learning greatly relies on the large-scale high-quality labeled training data. However, massive data annotation is extremely expensive and time-consuming. To improve the annotation efficiency and reduce the cost, engineers often collect such large-scale datasets from the crowdsourcing platforms~\cite{yan2014learning}, online queries~\cite{blum2003noise} and some image search engines~\cite{li2017learning}, which inevitably yield low-quality and noisy data.
%Even in some frequently-used and well-annotated datasets (ImageNet and QuickDraw, etal), there also contains portions of label-noisy data~\cite{northcutt2021pervasive}.
%The existence of the label noise can significantly degenerates the performance of the DNN, since the deep models can easily overfit to the noisy labels and construct artificial correlation between the features and the noisy labels. Thus, mitigating the side-effects of the noisy labels becomes a very crucial topic for large-scale representation learning.

Label-noise learning has drawn more and more attention in the deep learning community, e.g.,\cite{berthon2021confidence,yu2018learning,cheng2020learning,yang2021estimating,northcutt2017learning,han2018co}. The main reason is that accurately annotating large-scale datasets becomes extremely costly and sometimes even infeasible~\cite{karimi2020deep}. An effective way is to collect such large-scale datasets from the crowd-sourcing platform~\cite{yan2014learning} or online queries~\cite{blum2003noise}, which inevitably yield low-quality and noisy data. Thus, mitigating the side-effects of noisy labels becomes a very crucial topic. The noise model can be categorized as the \textit{class-conditional noise} (CCN) and the\textit{ instance-dependent noise} (IDN). In CCN, each instance from one class has a fixed probability of being assigned to another. While in IDN, the probability that an instance is mislabeled depends on both its class and features. In this paper, we focus on the more promising IDN approach, which considers a more general noise and can cope with real-world noise.

The traditional label-noise learning methods can be divided into two categories: algorithms with statistically inconsistent classifiers and algorithms with statistically consistent classifiers. In the first category, algorithms do not model the label noise distribution explicitly, they usually employ some heuristics to reduce the negative effects of the label noise~\cite{guo2018curriculumnet,han2018masking,han2018co,han2020sigua}. Although such approaches often empirically work well, the learned classifier from the data with label noise may not be statistically consistent and their reliability cannot be guaranteed.
To address this limitation, the classifier-consistent algorithms have been proposed.
Specifically, recent studies showed that estimating the transition matrix plays an important role in building  consistent classifiers for label-noise learning, as these methods can explicitly model the generation process of the noisy label~\cite{goldberger2016training,scott2013classification}.
%Then, classifier learned by only exploiting the noise data will converge to the optimal model trained with clean data as the number of training data goes to infinity~\cite{liu2015classification,patrini2017making,yu2018learning}.
However, it is very challenging to obtain the instance-dependent transition matrix (IDTM) for getting the noisy labels from the clean labels, because the IDTM $T(\textbf{x})$ as a function of instance $\textbf{x}$ is unidentifiable under IDN without any constraint.

\begin{figure*}[!t]
\centerline{\includegraphics[width=17.5cm]{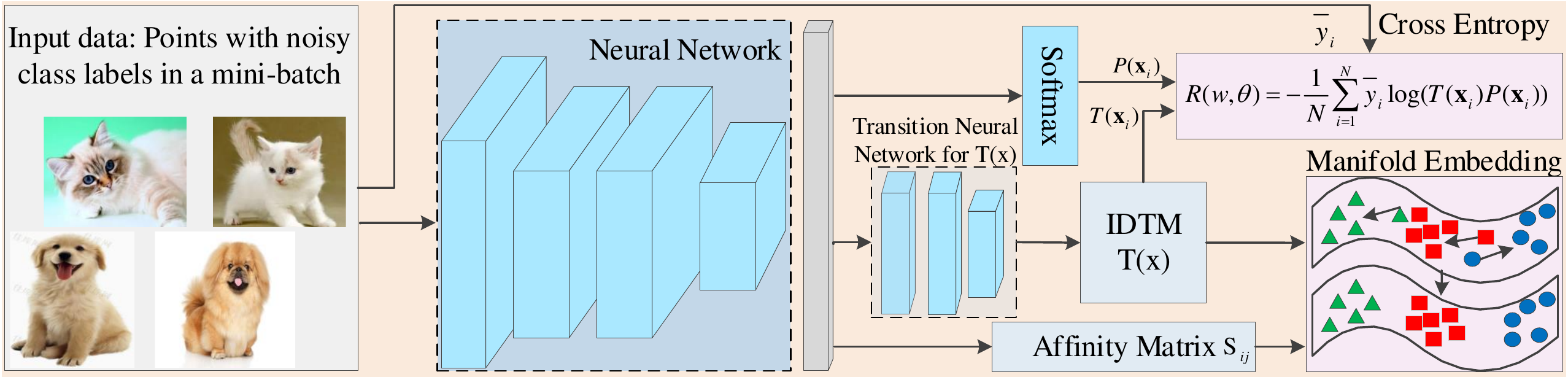}}
\caption{The proposed instance-dependent label-noise learning framework. We train a classifier in a statistically consistent manner through the proposed IDTM $T(\textbf{x})$, where $T(\textbf{x}_i) \in R^{K\times K}$ is estimated by the transition neural network. It is regularized by the manifold embedding to reduce the degree of freedom of $T(\textbf{x})$ and make it estimable in practice. In the manifold embedding $\mathcal{L}$, the affinity matrix $S_{ij}$ is obtained by finding the $k$-nearest neighbors in the instance feature space. Finally, we use the cross-entropy to train the classifier assisted by $T(\textbf{x})$. }
\label{fig1}
\vspace{-4mm}
\end{figure*}

Existing methods have tried to deal with this challenging and ill-posed problem from two perspectives. First, they have simplified the complex problem of estimating a matrix-valued function $T(\textbf{x})$ for the general label noise into a problem of estimating $T$ (i.e., a fixed matrix), which is known as CCN. Then, some anchor points (training data that certainly belong to some specific classes) are adopted to easily estimate the transition matrix $T$~\cite{liu2015classification,patrini2017making}. Although such methods have theoretical guarantees and have achieved success under some synthetic noisy labels or specific conditions, they are unable to cope with general real-world noise, i.e., IDN.
Second, several pioneer works considered strong assumptions and focused on how to simplify $T(\textbf{x})$ and significantly reduce its degree of freedom or complexity. For example, in part-dependent noise~\cite{xia2020part}, it is assumed that $T(\textbf{x})$ is a convex combination of a predefined number of fixed transition matrices, and their coefficients come from non-negative matrix factorization. In such a way, the estimation problem becomes parametric, and the degree of freedom of $T(\textbf{x})$ can be reduced. The issue of those methods is that simplifying the form of $T(\textbf{x})$ too much will certainly cause a large approximation error.

To address the problem of estimating IDTM $T(\textbf{x})$ under IDN, in this paper, we will not put any strong assumption on the form of $T(\textbf{x})$, but we will instead put some assumption on the \emph{geometry} of $T(\textbf{x})$. More specifically, we have noticed that, there are psychological and physiological evidences~\cite{logothetis1996visual,cohen2020separability,palmer1977hierarchical,shi2016improving} showing that we humans are more likely to annotate instances of similar appearances to the same class, and thus poor-quality or ambiguous instances of similar appearances are susceptible to be mislabeled to correlated or same noisy classes. Therefore, according to the basic principle that the noisy class-posterior probability $P(\bar{Y}=j|X=\textbf{x})$ can be inferred by the latent clean class-posterior probability $P(Y=i|X=\textbf{x})$ and IDTM $T(\textbf{x})$, we propose an assumption that ``the closer two instances are, the more similar their corresponding transition matrices will be''. This can be interpreted as that the instance adjacent relationships of one category in the feature space should be consistent with those in the transition matrix space.

Motivated by this practically useful assumption, we propose to estimate  $T(\textbf{x})$ by formulating the assumption into the \emph{manifold embedding} as shown in Figure~\ref{fig1}. Specifically, we make use of the manifold assumption, and require that if $\textbf{x}_i$ and $\textbf{x}_j$ are close in the feature space, then $T(\textbf{x}_i)$ and $T(\textbf{x}_j)$ should also be close (in terms of a matrix norm).
%The manifold embedding objective enforces the following properties for the learned transition matrices: 1)Similar instances within one noisy category should correspond to similar or even identical transition matrix, specially for some very confident clean examples, they should have identical fixed transition matrix; 2) Transition matrices corresponding to instances from different categories should be far apart and have large margin.
%Specifically, to further improve the effectiveness of the proposed method, we also considered the kernel trick to define the adjacent matrix for learning the IDTM.
Going along this line, though we do not reduce the complexity of $T(\textbf{x})$ directly since we do not further simplify it, we still effectively reduce the degree of freedom of the linear system $P(\bar{y}_i|\textbf{x}_i)=T(\textbf{x}_i)P(y_i|\textbf{x}_i), i=1,..,N$ and make $T(\textbf{x})$ stably estimable in practice. Here, $T(\textbf{x})$ can be regarded as practically stable, since adding such a smoothness assumption stops $T(\textbf{x})$ from changing too much in a tiny neighborhood and then it should be Lipschitz continuous. Thus, it should be uniquely determined given infinite data or the underlying data distribution.
%We think this manifold-regularized estimation technique works by greatly reducing the estimation error without hurting much approximation error about the estimation problem of $T(\textbf{x})$.
Finally, we conduct extensive experiments on various datasets, which illustrates the proposed method is superior to state-of-the-art approaches for label-noise learning under IDN.

%To fulfil the proposed assumption, we creatively formulate the assumption into the manifold embedding framework, to accurately estimate the instance-dependent transition matrix for noisy label learning. The manifold learning keeps the manifold distribution consistency between two different feature spaces~\cite{cayton2005algorithms,lin2008riemannian,shi2016improving}. That is to say, the manifold of the learned transition matrices is identical to that of the instances in the feature space. Specifically, in the proposed noisy-label learning framework, we design manifold regularization from the traditional one to the kernel-wise perspectives. Finally, we did extensive experiments on various datasets, which illustrate the effectiveness of the proposed method and prove that the proposed hypothesis is very useful for the instance-dependent noisy-label learning.

The main contributions are summarized as follows:
\begin{itemize}
%\item We are the first to propose the practical assumption that ``the closer the two instances are, the more similar of their corresponding transition matrices will be'', which is consistent with human's psychological habits.
%\item We are the first to propose a smoothness assumption on the \emph{geometry} of $T(\textbf{x})$ to effectively reduce its degree of freedom, and make $T\textbf{x}$ stably estimable.
\item We are the first to propose the practical assumption on the \emph{geometry} of $T(\textbf{x})$ that ``the closer two instances are, the more similar their corresponding transition matrices should be'', which aims to reduce the degree of freedom of $T(\textbf{x})$, and make it stably estimable in practice.

%\item To fulfill the proposed hypothesis, we formulate the assumption into the manifold embedding formulation. This allows us to keep the instance adjacent relationships of one category in the features space to be consistent with those in the transition matrix space, meanwhile push apart the transition matrices corresponding to instances from different categories.
\item We formulate the assumption into manifold embedding, which allows to keep the instance adjacent relationships in the features space to be consistent with those in the transition matrix space. By this way, the proposed manifold-regularized method can greatly reduce the estimation error without hurting much approximation error about the estimation problem of $T(\textbf{x})$.

    %. Besides, the manifold regularization is designed from the traditional and kernel-wise perspectives.

\item Extensive experiments on various datasets demonstrate superior classification performances over current state-of-the-art methods on both synthetic IDN datasets (MNIST, CIFAR10, CIFAR100) and two real-world noisy datasets (Clothing1M and Food101N).
\end{itemize}

%\vspace{-5mm}
\section{Related Works}

\textbf{Typical label-noise models} can be categorized as the random classification noise (RCN) model, the CCN model, and the IDN model. In the RCN model, the clean labels flip randomly with a fixed noise rate $\rho\in [0,1/2)$~\cite{kearns1998efficient,angluin1988learning}.
The CCN model is a nature extension of the RCN model for multi-class classification~\cite{stempfel2009learning, patrini2016loss,ma2018dimensionality}. It assumes that the flip rate for each instance from class $i$ to class $j$ depends on the latent clean class. Thus, it is possible to model some similarity information between classes. For example, we expect that the image of a ``dog'' is more likely to be miss-labeled as "cat" than "boat"\cite{berthon2021confidence}. Common methods to handle the CCN model include the ``loss correction" ~\cite{patrini2017making,ma2018dimensionality} and the ``label correction'' methods~\cite{tanaka2018joint,reed2014training}. The IDN model considers more general case of label noise, where the probability of an instance being mislabeled depends on the features and class of the instance itself. Intuitively, IDN is quite realistic and applicable, as the poor-quality and ambiguous instances are more prone to be labeled wrongly in real-world datasets. However, it is very complex to model IDN without any additional assumption. Our work in this paper, aims to estimate this realistic IDN model by considering practical useful assumptions on the \emph{geometry} of $T(\textbf{x})$,  which aims to reduce the degree of freedom of $T(\textbf{x})$, and make it stably estimable in practice.
%which is shown to be consistent with human's psychological habits~\cite{logothetis1996visual,cohen2020separability,palmer1977hierarchical} .

\textbf{Estimating the transition matrix} is one popular direction to build statistically consistent classifier. It helps to significantly reduce the side-effects of noisy labels, by inferring clean distribution based on the transition matrix and the noisy class-posterior probabilities, statistically. We first review representative works under the CCN. By leveraging the class-dependent transition matrix (CDTM) $T$, the training loss on the noisy data can be corrected. There exist many algorithms to estimate the CDTM $T$~\cite{hendrycks2018using,shu2020meta,yao2020dual,liu2015classification,li2021provably,xia2019anchor}. For example, Liu~\etal~\cite{liu2015classification} introduced the anchor points assumption to estimate $T$, Li~\etal~\cite{li2021provably} tried to optimize the transition matrix by minimizing the volume of $T$.
To estimate the IDTM $T(\textbf{x})$, existing works rely on various assumptions, for example, Cheng~\etal~\cite{cheng2020learning} proposed to estimate $T(\textbf{x})$ with a bounded noise rate, Xia~\etal~\cite{xia2020part} proposed to approximate the IDTM $T(\textbf{x})$  by utilizing the part regularization of the transition matrices.
Berthon~\etal~\cite{berthon2021confidence} introduced the instance-level forward correction method to estimate $T(\textbf{x})$, Yang~\etal~\cite{yang2021estimating} proposed to infer the Bayes optimal distribution instead of the clean distribution. Although above advanced methods achieved success empirically, some strong assumptions limit their applications in practice~\cite{yang2021estimating}. In contrast, our work proposes a manifold-regularized method to reduce the estimation error without hurting much approximation error of $T(\textbf{x})$, which achieves superior performances for label-noise learning.

\textbf{Extracting confident clean examples} is crucial for optimizing the transition matrix $T(\textbf{x})$ when only noisy data is available. To accurately estimate the transition matrix, we usually require that some clean data of each class are given. When clean data is not available, they are required to be extracted from the noisy data automatically, to construct confident clean dataset for optimizing $T(\textbf{x})$. The current effective methods mainly include but not limited to the following approaches: the distillation method~\cite{yang2021estimating}, sample sieve approach~\cite{cheng2020learning,lyu2019curriculum}, loss distribution modeling by a Gaussian mixture model~\cite{li2020dividemix}, confidence-based sample collection~\cite{berthon2021confidence,han2020sigua}, small-loss-based methods~\cite{wang2019co,han2018co}, and some early stopping techniques~\cite{bai2021understanding}. When the confident clean examples are obtained, the IDTM $T(\textbf{x})$ for each instance can be learned. Besides, there also exist many other semi-supervised learning methods~\cite{li2020dividemix,nguyen2019self,cheng2020learning,guo2018curriculumnet,han2018masking}, which transform the label-noise learning into the semi-supervised learning by using the extracted confident clean examples. In this paper, we also need to adopt such a method to extract confident clean examples for optimizing the IDTM $T(\textbf{x})$.

%\textbf{Other Representative Works} include but not limited to the following methods: the sample sieve based methods~\cite{cheng2020learning,lyu2019curriculum,han2018co}, which progressively sieves out corrupted examples to deal with the instance-dependent noise; the newly proposed robust loss functions~\cite{liu2020peer,xu2019l_dmi,zhang2017mixup}, which can be promoted as a robust candidate loss when facing potential noisy labels; and some semi-supervised learning technics~\cite{li2020dividemix,nguyen2019self}, which transform the noisy label learning into the semi-supervised learning by using the filtered instances.

 \vspace{-2mm}
\section{Label-Noise Learning Method} \label{RelatedWork}
%-------------------------------------------------------------------------
%In this section, we propose a new perspective to estimate the IDTM $T(\textbf{x})$. The core idea is that we utilize the manifold embedding objective to fulfil the newly proposed practical assumption, ``the closer distance between the similar instances in one category, the more similar of their corresponding transition matrix'', and the important observations for the IDTM  $T(\textbf{x})$. Specifically, we firstly introduce the problem settings and some important definitions,
%and then we collect a distilled dataset with theoretically guaranteed latent clean labels out of the noisy training data,
%and then we described constraint criterion of constructing the adjacent matrix for manifold learning, finally we jointly train a deep neural network to learn the transition matrix $T(\textbf{x})$ for each instance and the statistically consistent classifier on the given noisy dataset.
In this section, we obtain a statistically consistent classifier through the estimated IDTM $T(\textbf{x})$. Specifically, as illustrated in Figure~\ref{fig1}, the proposed method mainly consists of the following components: the input noisy data, confident clean example-extracting module, where we have cast them as a whole in Figure~\ref{fig1}, the backbone network and the transition neural network aiming to learn the instance features and estimate $T(\textbf{x})$ respectively, the cross-entropy loss training with noisy labels, and the proposed manifold-regularized objective. Finally, we jointly train the DNN to learn  $T(\textbf{x})$ for each instance, and obtain a consistent classifier $f(\textbf{x}; \textbf{w})$ on the given noisy data.
% Instance-Dependent with Manifold-Regularized
%on the given noisy data.

 \vspace{-1mm}
\subsection{Problem Setting}
Define $D$ be the distribution of pair-wise random variables $(\textbf{X},\textbf{Y}) \in \mathcal{X}\times\mathcal{Y}$, where $\textbf{X}$ denotes the variable of training samples, $\textbf{Y}$ is the variable of corresponding labels, $\mathcal{X}\in \mathbb{R}^{d}$ and $d$ is the instance feature dimension, $\mathcal{Y} = \{1,2,\ldots,K\}$ and $K$ is the total number of the label classes. The classification problem is to predict the label $y \in \mathcal{Y}$ for each  given instance $\textbf{x}\in \mathcal{X}$.
However, in some real-world classification problems, it is not easy or even infeasible to directly obtain large-scale training samples independently from the clean distribution $D$, since the clean labels are often randomly corrupted into the noisy labels when being observed.
%However, in many real-world classification problems, training examples independently drawn from the clean distribution $D$ are unavailable. Before being observed, the clean labels are often randomly corrupted into the noisy labels.

Define $\bar{D}$ be the distribution of these noisy examples $(\textbf{X},\bar{\textbf{Y}}) \in \mathcal{X}\times \bar{\mathcal{Y}}$, where $\bar{\textbf{Y}}$ denotes the random variable of noisy labels. This paper targets at the classification problem when we can only access a set of $N$ training examples with IDN denoted by $\bar{\mathcal{D}}:=\{(\textbf{x}_i,\bar{y}_i)\}_{i=1}^{N}$, where examples $(\textbf{x}_i,\bar{y}_i)$ is independently drawn according to $\bar{D}$.
%Our method aims to learn a robust classifier with these noisy training samples that can accurately classify the test instances.

\textbf{The IDTM} $T(\textbf{x})$ is defined to build the bridge between the clean distribution $D$ and the noisy distribution $\bar{D}$. As described in Eq.~(\ref{Eq:Infer}), the noisy class-posterior probability $P(\bar{Y}|\textbf{X})$ can be inferred by the IDTM $T(\textbf{x})$ and the clean class-posterior probability $P(Y|\textbf{X})$,  where $T(\textbf{x})=(T_{i,j}(\textbf{x}))_{i,j=1}^K \in [0,1]^{K\times K}$.

\begin{equation}\label{Eq:Infer}
  P(\bar{Y}=j|\textbf{X}=\textbf{x})=\sum_{i=1}^{K}T_{ij}(\textbf{x})P(Y=i|\textbf{X}=\textbf{x}),
\end{equation}
where the IDTM is defined as $T_{ij}(\textbf{x})=P(\bar{Y}=j|Y=i,\textbf{x})$. We can clearly see that $T(\textbf{x})$ depends on the actual instance, and it is tremendously complex as the noise is now characterized by $K^2$ functions over the instance feature space $\mathcal{X}$, which can be very high-dimensional.

%In many real-world classification problems, collecting clean labeled examples independently from $D$ is unavailable. Noisy labels  inevitably occur in many datasets.
%For the classification tasks with noisy label, we hope to train a statistically consistent classifier while having only access to the training samples from the noisy distribution $\bar{D}$ of random variable $(\textbf{X},\bar{\textbf{Y}}) \in \mathcal{X}\times\bar{\mathcal{Y}}$. Given a training dataset with Instance Dependent Noise (IDN) $\bar{\textbf{S}}=\{\textbf{x}_i, \bar{y}_i\}_{i=1}^n$, which are drawn from the noisy distribution $\bar{D}$, the noisy class posterior $P(\bar{Y}|\textbf{x})$ can be inferred by using the transition matrix $T(\textbf{x})$ and the latent clean class posterior $P(Y|\textbf{x})$, where $T(x)=(T_{i,j}(\textbf{x}))_{i,j=1}^K \in [0,1]^{K\times K}$:
%\begin{equation}\label{Eq:Infer}
%  P(\bar{Y}=j|\textbf{X}=\textbf{x})=\sum_{i=1}^{K}T_{i,j}(\textbf{x})P(Y=i|\textbf{X}=\textbf{x}),
%\end{equation}
%where the transition matrix function is defined as $T_{i,j}(\textbf{x})=P(\bar{Y}=j|Y=i,\textbf{x})$. We can clearly see that that transition matrix $T(\textbf{x})$ depends on the actual instance, and it is tremendously complex as the noise is now characterized by $K^2$ functions over the latent space $\textbf{X}$.

Therefore, the goal is to obtain a reliable classifier from the noisy training dataset that could classify the test instances accurately, by accurately estimating the IDTM $T(\textbf{x})$. Specifically, in Eq.~(\ref{Eq:Infer}), only the noisy class-posterior probability $P(\bar{Y}|\textbf{X})$ can be obtained by exploiting the noisy data. To accurately estimate $T(\textbf{x})$, there contain two important steps: 1) extracting confident clean examples; 2) optimizing the IDTM $T(\textbf{x})$ based on the given noisy labels and the extracted confident clean examples.

\textbf{Extracting confident clean examples} is very crucial for optimizing the IDTM $T(\textbf{x})$ as can be inferred from Eq.~(\ref{Eq:Infer}). In the related work section, we have listed many useful sample-sieve approaches.
In this paper, we adopt the example distillation method~\cite{yang2021estimating} to extract confident clean examples. This method can extract a distilled sub-dataset with theoretically guaranteed Bayes optimal labels out of the noisy dataset.
%We have set the noise rate upper bound parameter $\rho_{\mathrm{max}}$ as 0.3 in our experiments, which is the same as that in~\cite{yang2021estimating}.
Note that our method is not limited to the above distillation-based example extraction method, but many other sample-sieve approaches can also be used. Then, we can train a DNN on the extracted sub-dataset to learn the transition matrix $T(\textbf{x})$ to model the relationships between the confident clean data distribution $D$ and the noisy data distribution $\bar{D}$. Here, we denote as $\bar{\mathcal{D}}^{\mathrm{s}}:=\{(\textbf{x}_i^{\mathrm{s}},\bar{y}_i)\}_{i=1}^{N^{\mathrm{s}}}$ the extracted examples. For simplicity, we still use $\bar{\mathcal{D}}:=\{(\textbf{x}_i,\bar{y}_i)\}_{i=1}^{N}$ in the following descriptions.

%\subsection{Collecting Latent Clean Examples}
%In this subsections, we formally shows how to collect the confident clean examples automatically from the noisy dataset, which can be used for training the transition matrices $T(\textbf{x})$.
%To estimate the transition matrices, we usually require that anchors for each class are given. If anchor points are not available, they can be learned from the noisy data automatically to construct a latent clean dataset for optimizing the transition matrices $T(\textbf{x})$. The current effective methods mainly include but not limited to the following approaches: the distilled examples~\cite{yang2021estimating}, sample sieve approach~\cite{cheng2020learning,lyu2019curriculum}, loss distribution modeling by GMM~\cite{li2020dividemix}, early stopping technic based confident example mining~\cite{bai2021understanding}. In this paper, we adopt the the example distillation method as illustrated in ~\cite{yang2021estimating}, where it can collect a distilled dataset with theoretically guaranteed Bayes optimal labels out of the noisy dataset. Then, we can train a deep neural network on the collected distilled dataset to learn the transition matrices between the latent clean labels and the noisy labels.

\subsection{The Label-Noise Learning Framework}
%Given the sieved latent clean examples $(\textbf{x}^{sieved},\bar{y},\hat{y})$, where $\hat{y}$ denotes the latent label for the sieved clean example $\textbf{x}^{sieved}$, we train the transition network parameterized by $\theta$ to estimate the instance-dependent label-noise transition matrices $T(\textbf{x})$, which models the probability of observing a noisy label $\bar{y}$ from the given input image $\textbf{x}^{sieved}$ and its corresponding estimated latent clean label $\hat{y}$:
Given the extracted training examples $\{(\textbf{x}_i^{\mathrm{s}},\bar{y}_i)\}_{i=1}^{N^{\mathrm{s}}}$, we train the transition neural network parameterized by \textbf{$\theta$} to estimate $T(\textbf{x})$, which models the probability of observing a noisy label $\bar{y}$ from the given input instance $\textbf{x}$ and its corresponding estimated latent clean label $\hat{y}$.
%where $\hat{y}$ is different from the ground-truth label $y$ if the class posterior probability is wrong.
Then, it can be expressed as $T_{ij}(\textbf{x};\theta) = P(\bar{Y}=j|\hat{Y}=i,\textbf{X}=\textbf{x};\theta)$,
%\begin{equation}\label{Eq:T(x)}
%  T_{i,j}(\textbf{x}^{sieved},\theta) = P(\bar{Y}=j|\hat{Y}=i,\textbf{X}=\textbf{x}^{sieved};\theta),
%\end{equation}
where $T_{ij}(\textbf{x};\theta)\in \mathbb{R}^{K\times K}$, $\hat{Y}$ is the variable for the labels of the extracted confident clean data.
%which can be obtained by training a classification network $f(\cdot |\textbf{w})$ parameterized by $\textbf{w}$. Specifically, the class posterior probability for instance $\textbf{x}$ can be obtained by $\hat{y}=P(\hat{Y}|\textbf{x};\textbf{w}) = f(\textbf{x}|\textbf{w})$, where $f(\textbf{x}|\textbf{w}) \in \mathbb{R}^{K\times 1}$ is the output of the classifier.
To extract confident clean data, we need to learn the noisy class-posterior probability, which can be obtained by a classifier  $f(\bullet; \textbf{w})$ parameterised by $\textbf{w}$, i.e., $P(\hat{Y}|\textbf{x};\textbf{w}) = f(\textbf{x}; \textbf{w})$.

The probability of observing a noisy label $\bar{Y}$ given the input instance $\textbf{x}$ can be inferred as,
\begin{equation}\label{InferT}
  P(\bar{Y}=j|\textbf{x})= \sum_{i=1}^{K}P(\bar{Y}=j|\hat{Y}=i,\textbf{x})P(\hat{Y}=i|\textbf{x}).
\end{equation}
%\begin{equation}\label{InferT}
%  P(\bar{Y}=j|\textbf{x}; \textbf{w},\theta)&= \sum_{i=1}^{K}T_{i,j}(\textbf{x},\theta)P(\hat{Y}=i|\textbf{x};\textbf{w}),\\
%  &=\sum_{i=1}^{K}P(\bar{Y}=j|\hat{Y}=i,\textbf{x};\theta)P(\hat{Y}=i|\textbf{x};\textbf{w}),
%\end{equation}
Obviously, to infer the noisy label $\bar{Y}$ given the instance $\textbf{x}$, we need to optimize two sets of parameters $\{\textbf{w},\theta \}$, where $\textbf{w}$ is for the classifier $f(\textbf{x}; \textbf{w})$, $\theta$ is for the IDTM $T_{ij}(\textbf{x}; \theta) = P(\bar{Y}=j|\hat{Y}=i,\textbf{X}=\textbf{x};\theta)$. Traditionally, we could jointly optimize the parameters $(\textbf{w},\theta)$ by minimizing the empirical risk on the inferred noisy labels and the ground-truth noisy labels as follows:
\begin{equation}\label{EmpiricalRisk}
   \mathop{\min}_{\textbf{w},\theta}R(\textbf{w},\theta)=-\frac{1}{N}\sum_{i=1}^{N}\bar{y}_i\log(T(\textbf{x}_i;\theta) f(\textbf{x}_i;\textbf{w}) ),
\end{equation}
where $N$ is the number of instances in the extracted confident clean dataset. Intuitively, we can optimize Eq.~(\ref{EmpiricalRisk}) directly to obtain the parameter set $(\textbf{w},\theta)$. However, the transition matrix $T(\textbf{x};\theta)$ must be hard to learn without any assumption, the ultimate reason is that the degree of freedom of $T(\textbf{x};\theta)$ is too high, and the linear system $P(\bar{y}|\textbf{x}_i)=T(\textbf{x}_i) P(\hat{y}_i|\textbf{x}_i),i=1,...,N$, has the same number of equations and variables. Existing methods focused on how to simplify $T(\textbf{x})$ itself and significantly reduce its degree of freedom or complexity, which will certainly cause an approximation error. While in our work, we will not put any strong restrictions on the form of $T(\textbf{x})$, instead we will put a mild assumption on the \emph{geometry} of $T(\textbf{x})$ by using the manifold regularization.
%That is to say, the objective function in Eq.~\ref{EmpiricalRisk} is optimized on the sieved training dataset, which means that the transition matrix network is trained on a biased set, i.e., the set of the sieved examples. The network will generalize well to the non-sieved examples if they share the same pattern with the sieved examples which causes label noise. Recent studies~\cite{yang2021estimating,xia2020part} empirically verified that the patterns that cause label noise are commonly shared. Our empirical experiments further show that the transition network generalize well to unseen examples and can achieve superior classification performances. Note that, we optimize the empirical risk in Eq.~\ref{EmpiricalRisk} simultaneously in the training process.

\subsection{Manifold-Regularized Transition Matrix}
Manifold learning typically aims to retain intrinsic neighbouring structures in the underlying lower-dimensional feature space. The classical manifold learning techniques such as LLE~\cite{roweis2000nonlinear} and Isomap~\cite{tenenbaum2000global} estimate the local manifolds via justifiable assumptions. Therefore, we adopt the manifold embedding techniques to fulfill our proposed assumption ``the closer two instances are, the more similar their corresponding transition matrices will be'', to make the IDTM $T(\textbf{x})$ practically learnable. With the manifold regularization, though we do not reduce the complexity of $T(\textbf{x})$ directly since we do not further model it, we still effectively reduce the degree of freedom of the linear system $P(\bar{y}_i|\textbf{x}_i)=T(\textbf{x}_i)P(y_i|\textbf{x}_i)$ and make $T(\textbf{x})$ stably estimable. Meanwhile, $T(\textbf{x})$ can be regarded as practically stable, since adding such a smoothness assumption stops $T(\textbf{x})$ from changing too much. It should be uniquely determined given infinite data or the underlying data distribution.

%Besides, the manifold embedding regularisation can also help to reduce the complexity of the hypothesis space for the transition matrices, since the instance-dependent transition matrix $T(\textbf{x})$ is characterized by $C\times C$ functions over the sieved training examples as illustrated in Eq.~\ref{Eq:T(x)}, where $C$ is the relatively large in some situations. For example, we need to train 10,000 functions on the CIFAR100 datasets ($C=100$) to estimate $T(\textbf{x})$.
%As illustrated in Eq.~\ref{Eq:T(x)}, the instance-dependent transition matrix $T(\textbf{x})$ is characterized by $C\times C$ functions over the sieved training examples, where $C$ is the relatively large in some situations. For example, we need to train 10,000 functions on the CIFAR100 datasets ($C=100$) to estimate $T(\textbf{x})$. In order to reduce the complexity of the hypothesis space to make accurate estimation, we propose the practical assumption ``the closer the distance between two instances, the more similar of their corresponding transition matrix''. Then we fulfil the above assumption in the manifold embedding manner.~\cite{logothetis1996visual,cohen2020separability,palmer1977hierarchical,shi2016improving}

More specifically, we construct an intrinsic affinity graph to characterize the within-manifold consistency and the extrinsic affinity graph to characterize the between manifold relationships~\cite{shi2016improving}, respectively. %Meanwhile, since instances with different noisy labels corresponds to different effective row values in the transition matrix, then we also construct a penalty graph to characterize the margin between different manifolds.
The intrinsic graph is constructed by node adjacency relationships in all the manifolds, where each node is connected to its $k_1$-nearest neighbours within the same manifold. The extrinsic graph is constructed using the between-manifold  node adjacency relationships from different manifolds. We use the $k_2$-nearest neighbors between the $k$-th manifold and the other manifolds.

Specifically, the within-manifold regularization is to fulfil our assumption on the \emph{geometry} of $T(\textbf{x})$ that ``the closer two instances are, the more similar their corresponding transition matrices should be'', and can be expressed as:
\begin{equation}\label{InGraphLoss}
  \mathcal{M}_I=\sum_{i,j=1}^{N}S_{ij}^{(I)}||T(\textbf{x}_i)-T(\textbf{x}_j)||^2,
\end{equation}
\begin{equation}\label{InGraph}
S_{ij}^{(I)}=\left\{
\begin{array}{rcl}
1,        & {\textrm{if}\: \textbf{x}_j \in \mathcal{N}(\textbf{x}_i,k_1) \: \textmd{and}  \: \bar{y}_i=\bar{y}_j},\\
0,        & {\textrm{else},}
\end{array} \right.
\end{equation}
where $S_{ij}^{(I)}$ refers to element $(i,j)$ in the intrinsic affinity graph matrix $\textbf{S}^{I}=(S_{ij}^{(I)})_{N\times N}$, $T(\textbf{x}_i)$ indicates the IDTM for instance $\textbf{x}_i$,
$\mathcal{N}(\textbf{x}_i,k_1)$ indicates the $k_1$-nearest neighbours of the instance $\textbf{x}_i$, $\bar{y}_i=\bar{y}_j$ indicates that $\textbf{x}_i$ and $\textbf{x}_j$ are in the same manifold, the distance used for computing nearest neighbours is the Euclidean distance between $\textbf{x}_i$ and $\textbf{x}_j$ in the feature space. Obviously, minimizing the within-manifold consistency encourages the learned IDTM $T(\textbf{x})$ to be close, if their corresponding instances are close in the same category. This keeps the manifold in the instance feature space to be consistent with that in the transition matrix space.

Meanwhile, since instances with different noisy labels correspond to different effective row values in the transition matrix, we also construct a extrinsic graph to characterize the margin between manifolds. It can be expressed as:
\begin{equation}\label{BGraphLoss}
  \mathcal{M}_B=\sum_{i,j=1}^{N}S_{ij}^{(B)}||T(\textbf{x}_i)-T(\textbf{x}_j)||^2,
\end{equation}
\begin{equation}\label{BGraph}
S_{ij}^{(B)}=\left\{
\begin{array}{rcl}
1,        & {\textrm{if}\: \textbf{x}_j \in \mathcal{N}(\textbf{x}_i,k_2)\: \textmd{and}  \: \bar{y}_i\neq \bar{y}_j},\\
0,        & {\textrm{else},}
\end{array} \right.
\end{equation}
where $S_{ij}^{(B)}$ denotes element $(i,j)$ of the between-class affinity matrix $\textbf{S}^{B}=(S_{ij}^{(B)})_{N\times N}$, $\mathcal{N}(\textbf{x}_i,k_2)$ indicates the $k_2$-nearest neighbours of instance $\textbf{x}_i$, $\bar{y}_i\neq \bar{y}_j$ indicates $\textbf{x}_i$ and $\textbf{x}_j$ are from different manifolds.
%$\zeta_{k_2}(c_i)$ is the set of index pairs that are the $k_2$-nearestpairs among the set $(i,j)|i\in\chi_c,j\nsubseteq \chi_c$, and $\chi_c$ denotes the index set of the samples in the $c-th$ manifold.

Therefore, the overall proposed manifold-regularization on the IDTM $T(\textbf{x}_i;\theta)$ can be expressed as:
\begin{equation}\label{InGraphLoss}
  \mathcal{M}(\theta)=\mathcal{M}_I - \mathcal{M}_B.
\end{equation}
We can clearly see that, minimizing the manifold-regularization objective $\mathcal{M}(\theta)$ is equivalent to keep the manifold in the transition matrix space be consistent with that in the feature/label space. Thus, we fulfill the proposed practical useful assumption on the \emph{geometry} of $T(\textbf{x})$. %``the closer two instances are, the more similar their corresponding transition matrices will be''.

\vspace{1mm}
\textbf{Remarks}: The manifold embedding is usually used in the unsupervised or semi-supervised learning, its affinity matrix is constructed by the $k$-nearest neighbours in an unsupervised manner traditionally. However, in this work, we construct the affinity matrix for the manifold embedding while considering the given noisy labels as shown in Eq.~(\ref{InGraph}) and Eq.~(\ref{BGraph}). The main reason is that different instances with different given noisy labels correspond to different effective rows in their corresponding transition matrices $T(\textbf{x})$. We only used one row of $T(\textbf{x})$ to generate the noisy label. Then even two instances, $\textbf{x}_i$ and $\textbf{x}_j$, are very close in the feature space and have the same confident clean label, their corresponding transition matrices should be far apart if they have different given noisy labels. Therefore, we specially design the affinity matrix in the proposed manifold-regularized label-noise learning framework, as shown in Eq.(\ref{InGraph}), (\ref{BGraph}), (\ref{InGraphk}) and (\ref{BGraphk}), for optimizing the IDTM $T(\textbf{x})$.

\subsection{Kernel Trick to the Manifold Embedding}
To improve the effectiveness of the proposed manifold-regularization on the IDTM $T(\textbf{x}_i;\theta)$ , we further consider to adopt the kernel trick to pre-compute the affinity graph matrix~\cite{shi2016improving}. Concretely, $S_{ij}^{(I)}$ and $S_{ij}^{(B)}$ can be defined as,
\begin{equation}\label{InGraphk}
S_{ij}^{(I)}=\left\{
\begin{array}{rcl}
e^{-\frac{||\textbf{x}_i-\textbf{x}_j||^2}{\sigma^2}},        & {\textrm{if}\: \textbf{x}_j \in \mathcal{N}(\textbf{x}_i,k_1)\: \textmd{and}  \: \bar{y}_i=\bar{y}_j,}\\
0,        & {\textrm{else},}
\end{array} \right.
\end{equation}
\begin{equation}\label{BGraphk}
S_{ij}^{(B)}=\left\{
\begin{array}{rcl}
e^{-\frac{||\textbf{x}_i-\textbf{x}_j||^2}{\sigma^2}},&\: {\textrm{if}\: \textbf{x}_j \in \mathcal{N}(\textbf{x}_i,k_2)\: \textmd{and}  \: \bar{y}_i\neq \bar{y}_j,}\\
0,        & {\textrm{else},}
\end{array} \right.
\end{equation}
where $\sigma$ is one hyper-parameter to adjust the weight distribution in the affinity graph matrix.

%Note that, $S_{ij}^{(I)}$ and $S_{ij}^{(B)}$ are pre-computed in every training iteration, they don't participate in the gradient back-propagation during model optimization, though they contain some variables $\textbf{x}_i$ and $\textbf{x}_j$.

\subsection{Overall Objective Function}

Finally, the overall objective function can be expressed as Eq.(\ref{OverallLossFunction}),
\begin{equation}\label{OverallLossFunction}
  \mathop{\min}_{\textbf{w},\theta}\mathcal{L}(\textbf{w},\theta)=R(\textbf{w},\theta)+\lambda \mathcal{M}(\theta),
\end{equation}
where $\lambda$ is the hyper-parameter to balance the cross-entropy loss and the manifold embedding regularization.

\subsection{Optimization}
During optimization, the traditional back-propagation method (e.g., SGD) is used to learn the classifier $f(\textbf{x};\textbf{w})$ and the IDTM $T(\textbf{x})$. Therefore, it is required to compute the gradient of the objective function with respect to the output of the corresponding layers.

Define $\textbf{S}= (S_{ij})_{N\times N} = \textbf{S}^{I}-\textbf{S}^{B}$, then the manifold-regularized objective can be re-written as~\cite{shi2016improving},
\begin{equation}\label{ManifoldRegularization}
  \mathcal{M}=\sum_{i,j=0}^{N}S_{ij}||T(\textbf{x}_i)-T(\textbf{x}_j)||^2=2tr(\textbf{T} \mathbf{\Phi} \textbf{T}^T),
\end{equation}
where $\textbf{T}=[T(\textbf{x}_1),T(\textbf{x}_2),...,T(\textbf{x}_N)]$,  $T(\textbf{x}_i) \in \mathbb{R}^{K\times K}$ can be reshaped as $K^2\times 1$ dimension in $\textbf{T}$, $\mathbf{\Phi} = \textbf{D}-\textbf{S}$, $\textbf{D}=diag(d_{11},d_{22},...,d_{NN})$, $d_{ii}=\sum_{j=1,i\neq j}^NS_{ij}, i=1,2,...,N$, $\mathbf{\Phi}$ is the Laplacian matrix of $\textbf{S}$, and $tr(\cdot)$ denotes the trace of a matrix.

The gradient of $\mathcal{M}(\theta)$ with respect to $T(\textbf{x}_i)$ can be derived as~\cite{shi2016improving},
\begin{equation}\label{ManifoldRegularization}
  \frac{\partial\mathcal{M}}{\partial T(\textbf{x}_i)}=2\textbf{T}(\mathbf{\Phi} + \mathbf{\Phi}^T)_{(:,i)}=4\textbf{T} \mathbf{\Phi}_{(:,i)},
\end{equation}
where $\mathbf{\Phi}_{(:,i)}$ denotes the $i$-th column of matrix $\mathbf{\Phi}$.

Please note that, whether the affinity graph matrix $\textbf{S}$ is in the traditional form or the kernel-wise version, they are pre-computed based on current instance features in the mini-batch. They work as constant values which do not involve in the gradient back-propagation.

\begin{algorithm}
% \SetAlgoNoLine  %去掉之前的竖线
        \caption{Instance-dependent Label-Noise Learning Algorithm }~\label{algorithm}
        \KwIn{Noisy training dataset $\bar{\mathcal{D}}=\{\textbf{x}_i, \bar{y}_i\}_{i=1}^N$}
        \KwOut{The final classifier $f(\textbf{x};\textbf{w})$ and the transition matrix $T(\textbf{x}; \theta)$. }

        \textbf{Warmup}: Train the DNN on the noisy dataset $\bar{\mathcal{D}}$ with the early-stop strategy to obtain the initial classifier $f(\textbf{x};\textbf{w})$; \\
        \While{Number of training epoch $\leq$ Max-Epoch}{
            $\bullet$ Extract confident clean examples using example distillation method~\cite{yang2021estimating} with current classifier $f(\textbf{x};\textbf{w})$ to form the sub-dataset $\bar{\mathcal{D}}^s=\{\textbf{x}^s_i, \bar{y}_i\}_{i=1}^{N^s}$; \\

            $\bullet$ Input the extracted confident clean examples into the backbone network;\\

            $\bullet$ Compute the affinity graph matrix $S_{ij}^{(I)}$ and $S_{ij}^{(B)}$ based on current instance features according to Eq.~(\ref{InGraph}) and~(\ref{BGraph}) or Eq.~(\ref{InGraphk})~and~(\ref{BGraphk}); \\

            $\bullet$ Optimize the DNN based on the loss function shown in Eq.~(\ref{OverallLossFunction}) to obtain new classifier $f(\textbf{x}; \textbf{w})$ and transition matrix $T(\textbf{x}; \theta)$.

    }
\end{algorithm}

\vspace{-5mm}
\section{Experiments}
In this section, we first introduce the experiment setup including the dataset, noisy type and implementation details. Next, we compare the proposed method with the state-of-the-art methods on four synthetic and two real-world noisy datasets, then conduct an ablation study to analyze the experimental results and some useful hyper-parameters.
%conduct an ablation study to show that the proposed manifold embedding is benefit to accurately learn the transition matrices. Finally, we present and analyze the experimental results on four synthetic and two real-world noisy datasets to show the effectiveness of the proposed method.
\subsection{Experiment Setup}
\textbf{Datasets}. Extensive experiments are conducted to illustrate the effectiveness of our method, on four manually corrupted datasets (i.e., F-MNIST~\cite{xiao2017fashion},SVHN~\cite{netzer2011reading},CIFAR-10~\cite{krizhevsky2009learning}, CIFAR-100~\cite{krizhevsky2009learning}) and two real-world noisy datasets ( i.e., Clothing1M~\cite{xiao2015learning} and Food-101N~\cite{lee2018cleannet}). F-MNIST has $28\times 28$ gray scale images of 10 classes including 60K for training and 10K for testing. SVHN  contains 10 classes of images with $73,257$  for training and $26,032$ for testing. CIFAR-10 contains 10 classes and CIFAR-100 contain 100 classes, and both of them contain 50K training images  and 10K testing images of size 32$\times$ 32.
Clothing1M has 1M images with real-world noisy labels from 14 fashion classes for training and 10K test images with clean label, where the estimated noisy label rate is $38.46\%$. Food101N contains 101 food categories with 310k training images  and 55K clean images for testing, which is also a real-world noisy dataset and has about $19.66\%$ noisy labels in the training dataset.

\textbf{Noisy type}. For the manually corrupted datasets (i.e., F-MNIST, SVHN, CIFAR-10, and CIFAR-100), we adopt exactly the same strategy to generate the instance-dependent label noise as previous methods~\cite{cheng2020learning,xia2020part}. The basic idea is to randomly generate one vector for each class (K vectors for all the classes) and project each instance feature onto the K vectors. The noise label is generated by jointly considering its clean label and the projected results. We have set different noisy rate for all the datasets from $10\%$ to $50\%$ to evaluate all the methods.

\begin{table*}\centering
\caption{Comparison with state-of-the-art methods on F-MNIST and CIFAR-10 datasets. The mean and standard deviation computed over five runs are presented. ``IDN-xx$\%$'' means the noise rate is xx$\%$ and noise type is ``IDN''.}
  \vspace{-2mm}
\label{table:CIFAR-10}
\resizebox{\textwidth}{!}{
\begin{tabular}{@{}cc|ccccc|ccccc@{}}
\toprule
  & \multirow{3}{*}{Method} & \multicolumn{5}{c|}{F-MNIST} & \multicolumn{5}{c}{CIFAR-10}                           \\
  \cline{3-12}
  %&                                   & AWA2        & CUB         & SUN        & \multicolumn{3}{c}{AWA2} & \multicolumn{3}{c}{CUB} & \multicolumn{3}{c}{SUN} \\ \cline{3-14}
  &               & IDN-$10\%$         & IDN-$20\%$          & IDN-$30\%$         & IDN-$40\%$     & IDN-$50\%$     & IDN-$10\%$         & IDN-$20\%$          & IDN-$30\%$         & IDN-$40\%$     & IDN-$50\%$          \\ \hline
  & CE (baseline) & 87.73$\pm$ 1.25     &87.63$\pm$ 1.11    & 85.25$\pm$ 0.57    &75.00 $\pm$0.25   &65.42$\pm$ 1.59    &88.86$\pm$ 0.23 &86.93$\pm$  0.17  &82.42$\pm$ 0.44    &76.68$\pm$ 0.23   &58.93$\pm$  1.54   \\
  \hline
  & GCE~\cite{zhang2018generalized}      &90.24 $\pm$0.16    &88.71$\pm$ 0.17    &85.90$\pm$ 0.23    &76.78$\pm$ 0.37    &67.67$\pm$ 0.58    &90.82$\pm$ 0.05  &88.89$\pm$  0.08    &82.90$\pm$ 0.51  &74.18$\pm$  3.10    &58.93$\pm$ 2.67    \\
  & DMI~\cite{xu2019l_dmi}      &90.14$\pm$ 0.22    &88.13$\pm$ 0.47    &85.90$\pm$ 0.23    &76.22$\pm$ 0.71    &64.84$\pm$ 1.28    &91.43$\pm$ 0.18   &\textcolor{red}{\textbf{89.99}}$\pm$  0.15    &86.87$\pm$  0.34  &80.74$\pm$  0.44    &63.92$\pm$ 3.92      \\
  & Forward~\cite{patrini2017making}   &90.78$\pm$ 0.30   &89.01$\pm$ 0.44    &86.51$\pm$ 1.20   &78.17$\pm$ 0.32    &68.31$\pm$ 1.07    &\textcolor{red}{\textbf{91.71}}$\pm$  0.08   &89.62$\pm$ 0.14   &\textcolor{red}{\textbf{86.93}}$\pm$ 0.15  &80.29$\pm$ 0.27    &\textcolor{red}{\textbf{65.91}}$\pm$ 1.22     \\
  & CoTeaching~\cite{han2018co} &90.54$\pm$ 0.35 &88.53$\pm$ 0.09  &87.37$\pm$ 0.14    &78.36$\pm$ 0.82    &67.81$\pm$ 1.02    &90.80$\pm$ 0.05   &88.43$\pm$ 0.08    &86.40$\pm$ 0.41  &80.85$\pm$ 0.97    &62.63$\pm$  1.51      \\
  & CoTeaching++~\cite{yu2019does} &90.67$\pm$0.49 &88.52$\pm$ 0.44   &87.33$\pm$ 0.87    &79.85$\pm$ 1.03    &68.86$\pm$ 1.39    &91.47$\pm$ 0.59   &89.78$\pm$ 0.34    &85.72$\pm$ 0.35  &\textcolor{red}{\textbf{81.00}}$\pm$ 0.82    &61.46$\pm$  1.36      \\
  & JoCor~\cite{wei2020combating}      &\textcolor{red}{\textbf{91.48}}$\pm$ 0.11   &89.24$\pm$ 0.09   &86.50$\pm$ 0.10    &77.15$\pm$1.04    &67.85$\pm$0.84    &91.42$\pm$  0.11   &89.30$\pm$ 0.27    &85.54$\pm$  0.82  &80.87$\pm$  0.91    &64.11$\pm$  2.57      \\
  & PeerLoss~\cite{liu2020peer}    &90.76$\pm$0.41   &87.06$\pm$  0.74   &84.40$\pm$ 0.93    &73.95$\pm$ 2.37    &65.79$\pm$ 2.49    &90.89$\pm$  0.07   &89.21$\pm$  0.63   &85.70$\pm$  0.56  &78.51$\pm$  1.23    &59.08$\pm$  1.05      \\
  & TMDNN~\cite{yang2021estimating}      &91.33$\pm$ 0.27     &89.70$\pm$ 0.14   &87.63$\pm$ 1.28    &78.40$\pm$ 3.69    &66.55$\pm$ 7.52    &90.45$\pm$ 0.72  &88.14$\pm$  0.66    &84.55$\pm$0.48  &79.71$\pm$ 0.95    &63.33$\pm$  2.75      \\
  & PartT~\cite{xia2020part}      &91.27$\pm$ 0.38     &\textcolor{red}{\textbf{89.78}}$\pm$ 0.43   &\textcolor{red}{\textbf{88.30}}$\pm$ 0.51    &\textcolor{red}{\textbf{80.75}}$\pm$ 2.86    &\textcolor{red}{\textbf{72.22}}$\pm$ 4.22    &90.32$\pm$ 0.15  &89.33$\pm$  0.70    &85.33$\pm$1.86  &80.59$\pm$ 0.41    &64.58$\pm$  2.86     \\ \hline
  & MEIDTM (Ours)      &91.78$\pm$ 0.87    &90.49$\pm$ 0.35   &88.74$\pm$ 0.25   &84.21$\pm$ 0.52    &73.67$\pm$ 3.76    &92.17$\pm$ 0.21   &91.38$\pm$ 0.34    &87.68$\pm$ 0.26  &82.63$\pm$ 0.24   &72.17$\pm$ 1.51\\
  & kMEIDTM (Ours) &\textbf{91.96}$\pm$0.08  &\textbf{90.83}$\pm$ 0.05   &\textbf{89.61}$\pm$ 0.65    &\textbf{85.81}$\pm$  0.44    &\textbf{76.43}$\pm$ 4.88    &\textbf{92.91}$\pm$  0.07   &\textbf{92.26}$\pm$  0.25    &\textbf{90.73}$\pm$  0.34  &\textbf{85.94}$\pm$  0.92    &\textbf{73.77}$\pm$ 0.82      \\
 \bottomrule

\end{tabular}
}
\end{table*}

\begin{table*}\centering
\caption{Comparison with state-of-the-art methods on SVHN and CIFAR-100 datasets. The mean and standard deviation computed over five runs are presented. ``IDN-xx$\%$'' means the noise rate is xx$\%$ and noise type is ``IDN''.}
  \vspace{-2mm}
\label{table:CIFAR-100}
\resizebox{\textwidth}{!}{
\begin{tabular}{@{}cc|ccccc|ccccc@{}}
\toprule
  & \multirow{3}{*}{Method} & \multicolumn{5}{c|}{SVHN} & \multicolumn{5}{c}{CIFAR-100}                           \\
  \cline{3-12}
  %&                                   & AWA2        & CUB         & SUN        & \multicolumn{3}{c}{AWA2} & \multicolumn{3}{c}{CUB} & \multicolumn{3}{c}{SUN} \\ \cline{3-14}
  &               & IDN-$10\%$         & IDN-$20\%$          & IDN-$30\%$         & IDN-$40\%$     & IDN-$50\%$     & IDN-$10\%$         & IDN-$20\%$          & IDN-$30\%$         & IDN-$40\%$     & IDN-$50\%$          \\ \hline
  & CE (baseline) &90.47$\pm$0.27	&89.85$\pm$0.16	&86.31$\pm$0.79	&80.59$\pm$0.56	&64.93$\pm$2.03   &66.55$\pm$0.23 &63.94$\pm$0.51  &61.97$\pm$1.16 &58.70$\pm$0.56 &56.63$\pm$0.69  \\
  \hline
  & GCE~\cite{zhang2018generalized}      &90.82$\pm$0.12	&89.48$\pm$0.66	&86.92$\pm$0.24	&81.95$\pm$1.45	&63.20$\pm$2.75 &\textcolor{red}{\textbf{69.18}}$\pm$0.14	&\textcolor{red}{\textbf{68.35}}$\pm$0.33	&\textcolor{red}{\textbf{66.35}}$\pm$0.13	&62.09$\pm$0.09	&56.68$\pm$0.75
    \\
  & DMI~\cite{xu2019l_dmi}    &92.66$\pm$0.58	&91.88$\pm$0.42	&88.44$\pm$0.85	&82.27$\pm$1.54	&68.72$\pm$2.32  &67.06$\pm$0.46	&64.72$\pm$0.64	&62.8$\pm$1.46	&60.24$\pm$0.63	&56.52$\pm$1.18
        \\
  & Forward~\cite{patrini2017making}   &92.01$\pm$1.10	&90.67$\pm$0.27	&86.04$\pm$0.40	&83.18$\pm$0.95	&70.72$\pm$2.00    &67.81$\pm$0.48	&67.23$\pm$0.29	&65.42$\pm$0.63	&\textcolor{red}{\textbf{62.18}}$\pm$0.26	&58.61$\pm$0.44
     \\
  & CoTeaching~\cite{han2018co} &91.11$\pm$0.16	&90.88$\pm$0.17	&88.21$\pm$0.62	&86.46$\pm$1.33	&70.04$\pm$1.05   &67.91$\pm$0.34	&67.40$\pm$0.44	&64.13$\pm$0.43	&59.98$\pm$0.28	&57.48$\pm$0.740
      \\
  & CoTeaching++~\cite{yu2019does} &92.64$\pm$0.43	&91.59$\pm$0.43	&87.55$\pm$1.26	&87.69$\pm$1.06	&72.36$\pm$1.39   &68.67$\pm$0.25	&68.30$\pm$0.69	&65.77$\pm$0.30	&61.75$\pm$0.53	&\textcolor{red}{\textbf{57.94}}$\pm$0.15
    \\
  & JoCor~\cite{wei2020combating}      &93.52$\pm$0.47	&93.47$\pm$0.40	&89.47$\pm$1.04	&\textcolor{red}{\textbf{88.56}}$\pm$1.28	&73.70$\pm$1.92    &68.48$\pm$0.49	&67.87$\pm$0.80	&65.73$\pm$0.55	&61.64$\pm$0.54	&57.75$\pm$0.80
      \\
  & PeerLoss~\cite{liu2020peer}   &92.59$\pm$0.56	&91.67$\pm$0.72	&89.86$\pm$0.67	&85.44$\pm$0.97	&73.91$\pm$2.30   &65.64$\pm$1.07	&63.83$\pm$0.48	&61.64$\pm$0.67	&58.30$\pm$0.80	&55.41$\pm$0.28
     \\
  & TMDNN~\cite{yang2021estimating}  &95.51$\pm$0.13	&\textcolor{red}{\textbf{94.83}}$\pm$0.64	&92.43$\pm$0.91	&86.91$\pm$1.17	&76.53$\pm$2.15   &68.42$\pm$0.42	&66.62$\pm$0.85	&64.72$\pm$0.64	&59.38$\pm$0.65	&55.68$\pm$1.43
    \\
  &PartT~\cite{xia2020part}   &\textcolor{red}{\textbf{95.56}}$\pm$0.45	&94.19$\pm$0.20	&\textcolor{red}{\textbf{92.56}}$\pm$0.83	&88.13$\pm$1.56	&\textcolor{red}{\textbf{77.04}}$\pm$2.56   &67.33$\pm$0.33	&65.33$\pm$0.59	&64.56$\pm$1.55	&59.73$\pm$0.76	&56.80$\pm$1.32
     \\ \hline
  & MEIDTM (Ours)     &95.72$\pm$0.40 	&95.48$\pm$0.01 	&94.23$\pm$0.27 	&92.00$\pm$0.10 	&78.25$\pm$0.35    &68.19$\pm$0.32 	&67.21$\pm$0.38 	&66.06$\pm$0.77 	&62.34$\pm$0.18 	&57.69$\pm$0.51      \\
  &kMEIDTM (Ours) &\textbf{96.38}$\pm$0.07	&\textbf{95.66}$\pm$0.02	&\textbf{94.68}$\pm$0.17	&\textbf{92.20}$\pm$0.23	&\textbf{80.22}$\pm$2.00   &\textbf{69.88}$\pm$0.45	&\textbf{69.16}$\pm$0.16	&\textbf{66.76}$\pm$0.30	&\textbf{63.46}$\pm$0.48	&\textbf{59.18}$\pm$0.16
     \\
 \bottomrule

\end{tabular}
}
\vspace{-3mm}
\end{table*}

%\vspace{-1mm}
\subsection{Implementation Details}
For fair comparison, we conduct all experiments on NVIDIA GeForce RTX 3090, and all methods are implemented on the same PyTorch platform. The backbone network we used on F-MNIST dataset is ResNet-18, while ResNet-34 network is used on SVHN, CIFAR-10 and CIFAR-100 datasets. For the two real-world dataset (Clothing1M and Food101), we adopt the ResNet-50 network pre-trained on ImageNet as the backbone network. The transition neural network in the framework is implemented by one fully-connected layer, where the input is the instance features, and the number of output nodes is $K \times K$ where $K$ denotes the number of classes on each dataset. The obtained $T(\textbf{x})$ is normalized in each row. The $k$-nearest neighbor parameters in Eq.~(\ref{InGraph}) and Eq.~(\ref{BGraph}) are set as $k_1=k_2=7$, the hyper-parameter $\sigma$ in Eq.~(\ref{InGraphk}) and Eq.~(\ref{BGraphk}) is set to 1.1.  The optimization strategy we used is SGD  with momentum 0.9, weight decay $10^{-3}$, and batch size 128. The initial learning rate is set to $10^{-3}$, which is divided by $10$ every $20$ epochs. Firstly, we train the network on all the noisy data with the early stop techniques as warm-up, where we have trained 5, 10, 20, 1, and 1 epochs on F-MNIST, CIFAR-10, CIFAR-100, Clothing1M and Food101N datasets for warm-up, respectively. Then, we use the initial classifier to extract confident examples from the noisy datasets based on the distillation method~\cite{yang2021estimating}. The algorithm flowchart can refer to Alg.~\ref{algorithm}.
%Given the extracted confident clean examples, we train the whole network to obtain the transition matrix $T(\textbf{x})$ and the classifier $f(\textbf{x}; \theta)$.
%Our implementation is available at \url{http://github.com/noiselabellearning/manifold}.

\subsection{Comparison with State-of-the-art Methods}
We compare our method with the following 10 representative works: 1) CE, which trains the classification network with the standard cross-entropy loss on the original noisy dataset; 2) GCE~\cite{zhang2018generalized}, which uses the mean absolute error and the cross-entropy loss to jointly optimize the model on noisy datasets; 3) DMI~\cite{xu2019l_dmi}, which proposed a information-theoretic loss function to robustly train the deep model on the noisy dataset; 4) Forward~\cite{patrini2017making}, which utilizes a CDTM $T$ to correct the loss function; 5) Co-teaching~\cite{han2018co} and Co-teaching++~\cite{yu2019does} propose to train two deep neural networks simultaneously to handle label noise; 6) JoCor~\cite{wei2020combating} adopted a joint training method with co-regularization; 7) PeerLoss~\cite{liu2020peer},which does not require a prior specification of the noise rates; 8) TMDNN~\cite{yang2021estimating} and PartT~\cite{xia2020part} proposed to estimate the IDTM $T(\textbf{x})$ for IDN using DNNs.

\begin{table*}[t]
\centering
\caption{Classification accuracy ($\%$)  on the Clothing1M dataset. (*) indicates that the implementation is based on the authors' code.}\smallskip
\label{table:Clothing1M}
  \vspace{-2mm}
\resizebox{2.1\columnwidth}{!}{
\begin{tabular}{c|cccccccc }
\toprule
Methods &CE (Baseline) &GCE~\cite{zhang2018generalized} &SL~\cite{wang2019symmetric}   &Co-teaching~\cite{han2018co} &JointOpt~\cite{tanaka2018joint}  &$L_{DMI}$~\cite{xu2019l_dmi} &PTD-R-V~\cite{xia2020part}  &ERL~\cite{liu2020early} \\
   \hline
Accuracy\ &68.94  &69.75 &71.02 &69.21 &72.16 &72.46 &71.67 &72.87\\
\midrule
Methods &ForwardT~\cite{patrini2017making} &JoCor~\cite{wei2020combating} &CORES~\cite{cheng2020learning} &CAL~\cite{zhu2021second}  &DivideMix*~\cite{li2020dividemix} &MEIDTM(Ours) &kMEIDTM(Ours) &kMEIDTM (+DivideMix)  \\
\midrule
Accuracy &69.84 &70.30 &73.24  &74.17 &74.67 &73.05 &73.34 &\textbf{74.82} \\
\bottomrule
\end{tabular}
}
\vspace{-3.5mm}
\end{table*}

\begin{table}[t]
\centering
\caption{ Classification accuracy ($\%$)  on the Food101N dataset.}\smallskip
  \vspace{-2mm}
\label{table:Food101N}
\resizebox{0.78\columnwidth}{!}{
\begin{tabular}{c|c }
\toprule
 Methods  &Accuracy  \\
\hline
CE(Baseline)  &81.44 \\ \midrule
CleanNet$_{WHard}$(cvpr2018)~\cite{lee2018cleannet} &83.47 \\
CleanNet$_{WSoft}$(cvpr2018)~\cite{lee2018cleannet} &83.95 \\
DeepSelf(cvpr2019)~\cite{han2019deep} &85.11 \\
NoiseResist(cvpr2021)~\cite{liu2021noise} &84.70 \\ \hline
DivideMix(iclr2020)*~\cite{li2020dividemix} &84.39 \\ \hline
kMEIDTM(+DivideMix) (Ours) &\textbf{85.61}\\
\bottomrule
\end{tabular}}
\vspace{-5mm}
\end{table}

\textbf{Results on the synthetic noisy datasets}. Table~\ref{table:CIFAR-10},\ref{table:CIFAR-100},\ref{table:Food101N} and \ref{table:Clothing1M} report the classification accuracy on datasets of F-MNIST, SVHN, CIFAR-10 and CIFAR-100 under five different noise ratios, respectively. Each table includes 10 representative works on corresponding datasets. Our proposed method has two variants as described in the tables: one is the proposed method with kernel-trick affinity matrix as illustrated in Eq.~(\ref{InGraphk}) and (\ref{BGraphk}), which is our final version denoted as ``kMEIDTM''; another is the proposed method where the affinity matrix is build by Eq.~(\ref{InGraph}) and Eq.~(\ref{BGraph}), denoted as ``MEIDTM''. The baseline method is the standard cross-entropy loss trained on the noisy dataset, denoted as ``CE''.

Compared with the representative works, the proposed method achieves top performances on all the four synthesized datasets under five noise ratios. The evaluation results shown in the four tables  can be summarized as follows,
\begin{itemize}
  \item Compared with the best performances shown in previous representative methods, the proposed method kMEIDTM outperforms the former by a margin of 0.48\% to 7.67\%, and our method outperforms the baseline method ``CE'' by a large margin of 2.55\% to 17.29\%.
  \item The superiority of the proposed method is gradually revealed along with the noise rate increases. As shown in the four table, our method outperforms the second best method by an average margin of 0.64 and 1.04 under IDN-10\% and IDN-20\%, while 3.12\% and 4.69\% under IDN-40\% and IDN-50\%, which illustrate our method can handle extremely hard situation much better.
  \item Our kernel version  kMEIDTM outperforms MEIDTM in almost all situations, by a margin of 0.06\% to 2.76\%.
\end{itemize}

\textbf{Results on the real-world datasets}. Table~\ref{table:Clothing1M} and~\ref{table:Food101N} show the classification results on the real-world Clothing1M and Food101N datasets. It can be seen that the proposed method ``MEIDTM'' can improve the baseline method ``CE'' by a margin of $4.11\%$, and then the kernel-wise method ``kMEIDTM'' further improves the classification accuracy to $73.34\%$.

Since the proposed IDTM estimation method can work as a plug-and-play module, we then integrate this module into the representative work ``DivideMix''~\cite{li2020dividemix} to further illustrate its effectiveness, denoted as ``kMEIDTM(+DivideMix)''.  Experimental results show that the proposed transition matrix can further improve the method of ``DivideMix''~\cite{li2020dividemix} by $0.15\%$ and $1.22\%$ on Clothing1M and Food101N datasets respectively, which is superior to state-of-the-art methods.

\subsection{Ablation Study}
\begin{figure}
  \centering
  \includegraphics[width=7.8cm]{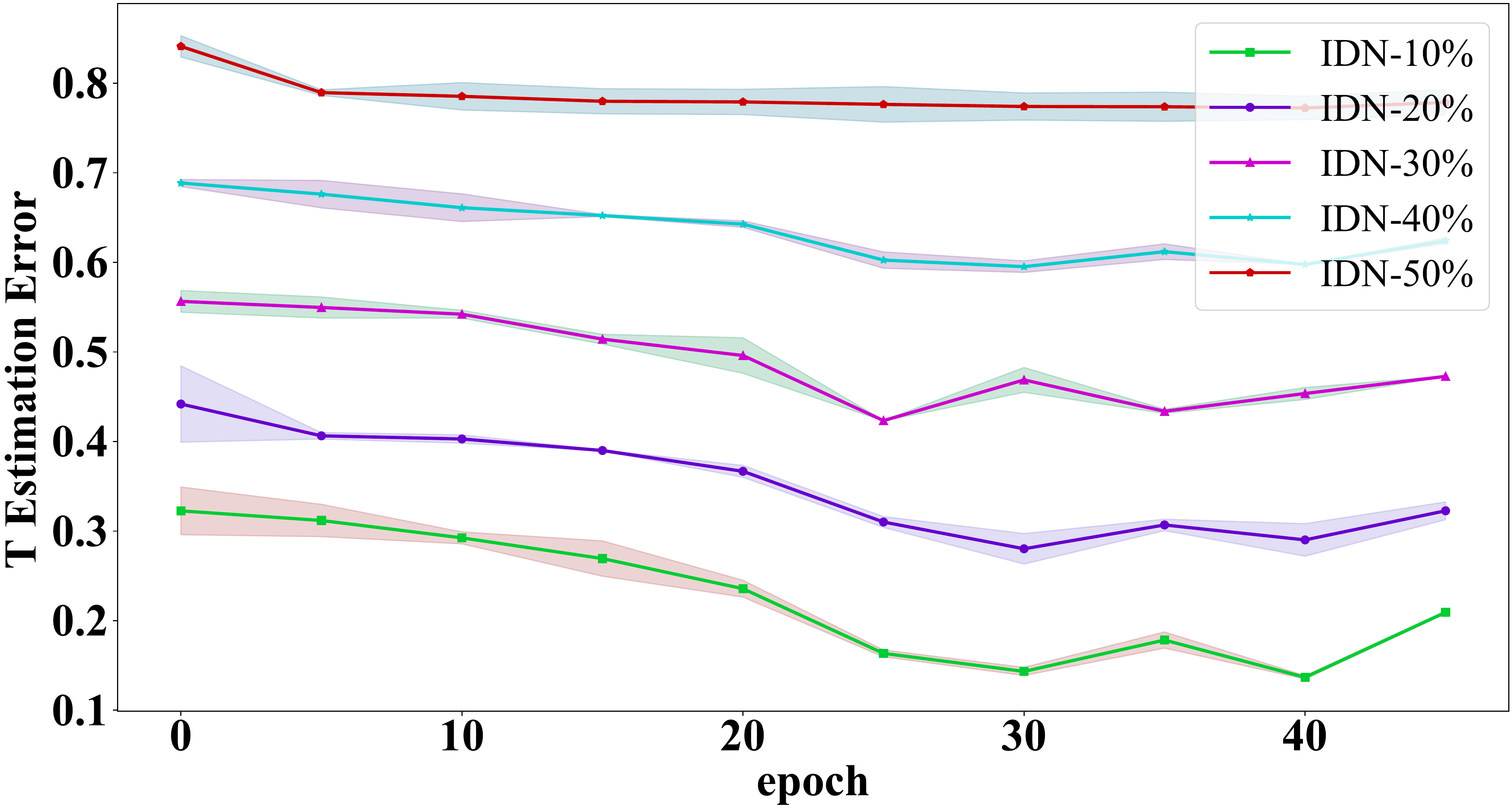}\\
  \vspace{-2mm}
  \caption{shows the transition matrix estimation error varying with the number of epoches during model training, under five different noise rates, on CIFAR-10 dataset.}\label{fig:TEstimationError}
  \vspace{-4mm}
\end{figure}

To evaluate the estimated IDTM $T(\textbf{x})$, we show the IDTM estimation error during model training under five different noise rates, on CIFAR-10 dataset, in Figure~\ref{fig:TEstimationError}. The error is measured by the $l_1$ norm between the ground-truth transition matrix and the estimated transition matrix. For each instance, we only analyze the estimation error of a specific low since the noisy is generated by one row of $T(\textbf{x})$. We summary Figure~\ref{fig:TEstimationError} as follows: 1) IDTM estimation error gets smaller and smaller during model training under five noise rates, which illustrate the effectiveness of the proposed method for $T(\textbf{x})$ optimization; 2) the lower noise rate, the better/easier estimation for $T(\textbf{x})$.
\begin{figure}
  \centering
  \includegraphics[width=7.8cm]{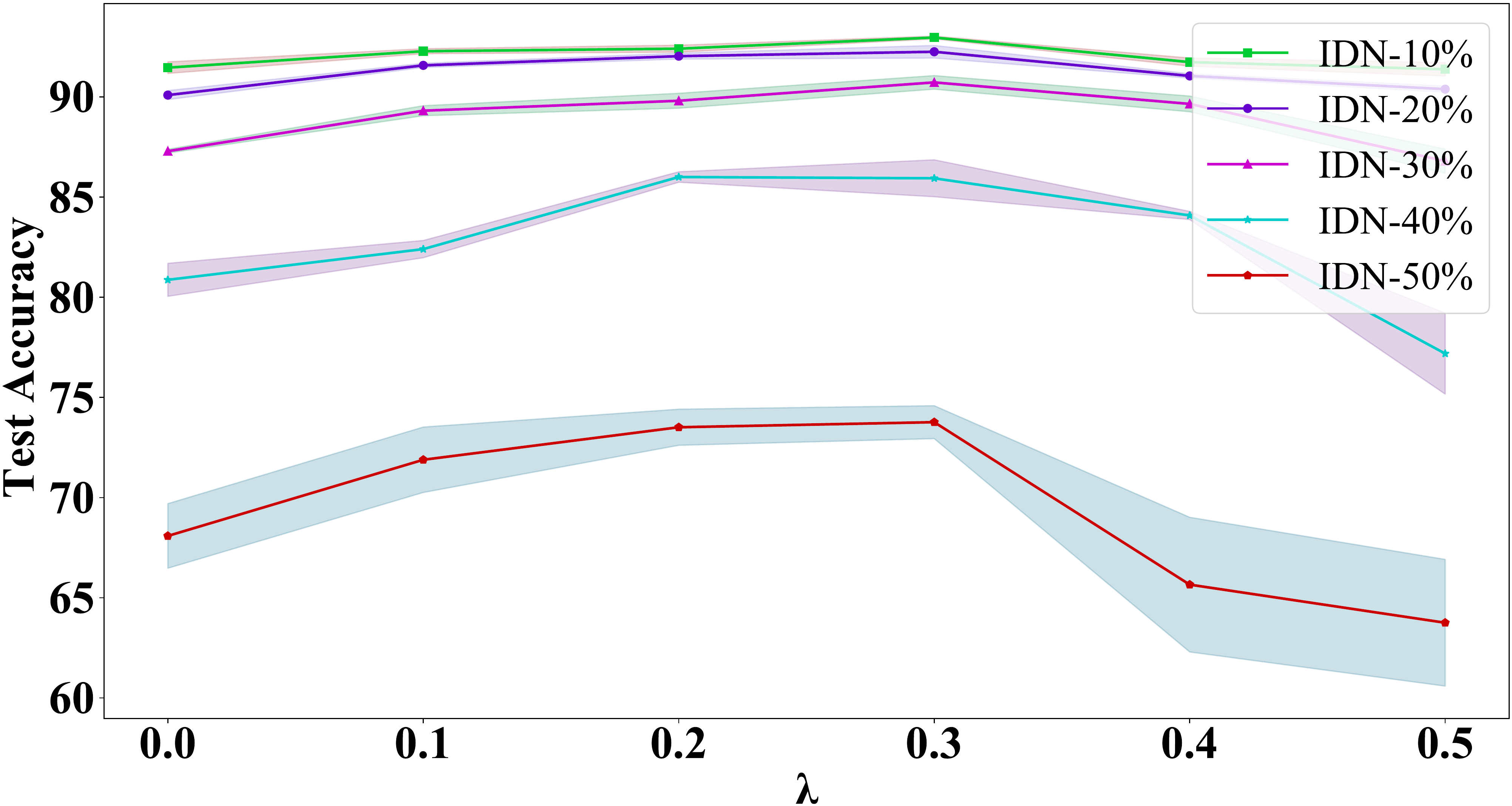}\\
  \vspace{-2mm}
  \caption{shows the classification accuracy varying with hyper-parameter $\lambda$ under five different noise rates on CIFAR-10 dataset.}\label{fig:LambdaAccuracy}
  \vspace{-4mm}
\end{figure}

To investigate the effect of hyper-parameter $\lambda$ on the model performance, we conduct experiments with various values of $\lambda$ on CIFAR-10 dataset under five noise rates, where each experiment is done with five runs. The results is shown in Figure~\ref{fig:LambdaAccuracy}. We can see that the test accuracy is not relatively sensitive to $\lambda$  under low noise rate, i.e., IDN-$10\%$, but it is sensitive under high noise rate, i.e., IDN-$50\%$. Overall, we can clearly see that our method yields the best performance when $\lambda$ is around 0.3. Based on this observation, we set $\lambda = 0.3$ in all our experimental evaluations.

\vspace{-2mm}
\section{Conclusion and Limitation}
In this paper, we focus on obtaining a consistent classifier under the challenging IDN. To address this problem, we propose the assumption on the \emph{geometry} of IDTM $T(\textbf{x})$ that ``the closer two instances are, the more similar their corresponding transition matrices should be''. Specifically, we formulate the assumption into the manifold embedding to effectively reduce the degree of freedom of $T(\textbf{x})$ and make it stably estimable. This method can directly reduce the estimation error without hurting much approximation error about the estimation problem of $T(\textbf{x})$. Extensive experimental results demonstrate the effectiveness of our method.
%  Besides, the proposed method can work as a plug-and-play module, which can help to further improve other methods.

\vspace{1mm}
\noindent \textbf{Limitation.} One major limitation in this study is that we just adopt one fully-connected layer with $ K\times K$ output nodes to work as the transition neural network (TNN) for learning $T(\textbf{x})$, which is a little bit simple. In the future, we will make deep analysis on the TNN design theoretically and practically, to learn robust classifier on the noisy dataset.
%the proposed method is based on the two-stage object detector, e.g., Faster-RCNN \cite{ren2016faster}, whose detection speed is relatively slow. We hope to integrate our method into some one-stage object detector,e.g., YOLOv5~\cite{jocher2021ultralytics}, to further improve the detection speed in the future.

\vspace{1mm}
\noindent \textbf{Acknowledgements:}
This work was supported in part by the National Key Research and Development Program of China under Grant 2018AAA0103202, in part by the National Natural Science Foundation of China under Grant 62176198, 61922066, 61876142, 62036007 and 62106184, in part by the Technology Innovation Leading Program of Shaanxi under Grant 2022QFY01-15, in part by Open Research Projects of Zhejiang Lab under Grant 2021KG0AB01 and in part by Australian Research Council Projects DE-190101473 and DP-220102121.

%%%%%%%%% REFERENCES
{\small
\bibliographystyle{ieee_fullname}
\bibliography{egbib}
}

\end{document}